# A hybrid deep-learning approach for complex biochemical named entity recognition

Jian Liu[1,3,10], Lei Gao[2,*], Sujie Guo[1,4], Rui Ding[1,5], Xin Huang[6], Long Ye[7,8], Qinghua Meng[8], Asef Nazari[9], and Dhananjay Thiruvady[9]

[1] HeFei University of Technology, Hefei, 230009
[2] CSIRO, Waite Campus, Urrbrae, SA 5064, Australia
[3] Intelligent Interconnected Systems Laboratory of Anhui Province, Hefei University of Technology
[4] Shanghai Engineering Center for Microsatellites, Shanghai 201203, China
[5] Xi'an Jiaotong University, Xi'an, 710049, China
[6] College of Computer and Information Engineering, Tianjin Normal University, Tianjin 300387, China
[7] School of Mechatronic Engineering and Automation, Shanghai University, Shanghai 200072, China
[8] Faculty of Business Information, Shanghai Business School, Shanghai 201400, China
[9] School of Information Technology, Deakin University, Geelong, Australia
[10] Anhui Province Key Laboratory of Industry Safety and Emergency Technology, Hefei 230601, Anhui, P.R. China

*Corresponding author:
Dr. Lei Gao
Senior Research Scientist
CSIRO Land and Water
Private Mail Bag 2, Waite Road
Glen Osmond SA 5064, Australia
Ph: +61-8-8273 8109
Fax: +61-8-8303 8750
Email: lei.gao@csiro.au

**Abstract**:

Named entity recognition (NER) of chemicals and drugs is a critical domain of information extraction in biochemical research. NER provides support for text mining in biochemical reactions, including entity relation extraction, attribute extraction, and metabolic response relationship extraction. However, the existence of complex naming characteristics in the biomedical field, such as polysemy and special characters, make the NER task very challenging. Here, we propose a hybrid deep learning approach to improve the recognition accuracy of NER. Specifically, our approach applies the Bidirectional Encoder Representations from Transformers (BERT) model to extract the underlying features of the text, learns a representation of the context of the text through Bi-directional Long Short-Term Memory (BILSTM), and incorporates the multi-head attention (MHATT) mechanism to extract chapter-level features. In this approach, the MHATT mechanism aims to improve the recognition accuracy of abbreviations to efficiently deal with the problem of inconsistency in full-text labels. Moreover, conditional random field (CRF) is used to label sequence tags because this probabilistic method does not need strict independence assumptions and can accommodate arbitrary context information. The experimental evaluation on a publicly-available dataset shows that the proposed hybrid approach achieves the best recognition performance; in particular, it substantially improves performance in recognizing abbreviations, polysemes, and low-frequency entities, compared with the state-of-the-art approaches. For instance, compared with the recognition accuracies for low-frequency entities produced by the BILSTM-CRF algorithm, those produced by the hybrid approach on two entity datasets (MULTIPLE and IDENTIFIER) have been increased by 80% and 21.69%, respectively.

# 1. Introduction

In recent years, artificial intelligence has helped increase the interactions among theoretical chemistry, computational chemistry, and synthetic chemistry. Deep neural networks have been recently used to analyze the rationality of chemical synthesis and find a large number of reverse synthetic routes in a short amount of time [1]. These computational tools make reaction analysis faster than manual approaches and allow efficient predictions of reactions of possible reagent combinations. There is no doubt that artificial intelligence, particularly deep learning, is revolutionizing our understanding of chemistry [2, 3].

Despite the advances, in the field of chemical drugs, there are still several important scientific activities and processes for the extraction of information that are done manually, taking plenty of experts' time. A number of these activities could be efficiently managed using **N**atural **L**anguage **P**rocessing (NLP) and other artificial intelligence tools [4, 5, 6]. As an example, the need for an intelligent tool that can automatically extract materials and chemical entities from the literature in the chemistry-related fields, is vital and urgent. The necessity of such tools provides the motivation to explore the **N**amed **E**ntity **R**ecognition (NER) technology [7].

NER refers to the identification of entities that have a specific meaning in the text, including names, names of places, and proper nouns [8, 9]. It was first presented as a concept and motivated as a research area at the Message Understanding Conference (MUC-6) in 1995 [10]. NLP, as an important tool in information extraction, helps the process of subsequent relationship extraction, event extraction, and disambiguation. To deal with NER, several methods including methods based on rules and dictionaries, statistical methods, and hybrid approaches have been used. Methods based on rules and dictionaries usually perform better when the task of recognition is in a specific corpus, such as the Dl-cotrain algorithm using decision tables proposed by Kwak et. al. [11]. However, rule-based methods are limited in their ability of effectively carrying out recognition tasks, since they rely on many complex rules. Moreover, they are context-sensitive, and require expert knowledge in specific fields and substantial efforts in maintaining rules and dictionaries. To overcome these drawbacks, the **H**idden **M**arkov **H**odel (HMM), the maximum entropy model (MaxEnt), the **C**onditional **R**andom **F**ield (CRF), and classical machine learning methods such as the support vector machines [12] have gradually replaced the aforementioned traditional methods.

The NER approaches based on machine learning models, consist of tasks that can be divided into pure artificial features, supervised tasks, semi-supervised tasks, and unsupervised tasks. Although traditional statistical methods based on artificial features have shown improvements in the field of data security, they suffer from deficiencies such as being computationally expensive, requiring large overheads for recognition, or poor generalizability. Supervised machine learning methods, such as the HMM named entity classifier, is based on a word similarity smoothing technique [13], and is characterized by high recognition accuracy. However, training data for these models are manually labeled, which tends to be a laborious task. Moreover, they need large training data, and this can be particularly problematic when the available data are scarce.

Similarly, semi-supervised and unsupervised techniques need very large amounts of training data. However, they have been very effective when dealing with low-frequency data. For example, Julian Brooke presented an NER system targeted specifically at fiction uses unlabeled data to obtain results [14], however, the approach clearly takes advantage of specific properties of (English) literature. The initial rule-based segmentation, for instance, depends on reliable capitalization of names, which is often not present in social media, or in most non-European languages. What is more, this approach cannot be successfully applied in cases where texts are relatively short.

With advances in deep learning and the large availability of computational resources, methods based on deep learning have demonstrated obvious advantages over traditional methods in NLP, and they have gradually become mainstream in the field of NER. They not only address some of the shortcomings of supervised machine learning methods (too computationally-intensive and time-consuming), but they also alleviate issues such as the lack of generalizability and large recognition workload commonly seen in machine learning methods based on artificial features.

NEC Labs America pioneered the idea of using deep learning for NLP [15]. Collobert et al. utilized embedding and multi-layer one-dimensional convolution structures for four typical NLP problems such as part-of-speech tagging. They used a combination of **C**onvolutional **N**eural **N**etworks (CNN) and CRF to achieve groundbreaking results in the optimization of the CONLL2003 corpus in the field of universal named entity recognition [16]. Subsequently, with the advent of long short-term memory (LSTM) and **Bi**directional **L**ong **S**hort-**T**erm **M**emory (BILSTM) based on **R**ecurrent **N**eural **N**etwork (RNN), the performance for NLP has gradually improved. The BILSTM-CRF model designed achieved an approximate F-value of 89% in the corpus [17]. Furthermore, the F-value of the CNN-LSTM model established by Chiu et al. exceeded 90% [18]. In 2016, Ma et al. proposed a method for end-to-end sequence tagging by BILSTM-CNNs-CRF, which led to the F-value of 91.21% for the CONLL2003 corpus [17].

Despite deep learning methods obtaining excellent performance in a generic NER, the performance of NER technologies in some specific fields such as chemistry, biology, finance, and so forth is still far from the ideal. Therefore, NER-related tasks in some fields need significantly more attention, as extracting biochemical entity information from text databases or scientific literature can assist interdisciplinary researchers in the field of biochemistry [19, 20].

There are two main difficulties in the identification of named entities in the field of biochemistry in comparison with generic NER. First, there has not yet been a unified naming methodology. There exist multiple expression methods for the same entity in addition to complicated and irregular naming issues including differences in English abbreviations, special characters, and so forth. Second, the number and types of biochemical entities are huge, and they are growing rapidly. At the moment, the number of new synthetic drugs is increasing exponentially, and new drugs are constantly being produced.

To deal with the challenges of NER in the field of biochemistry, Leaman et al. developed the tmChem system, which achieves an F-value of 87.395% on the CHEMDNER dataset [21]. In another work, Zheng used an attention-based recognition method to increase the F-value to 90.77% via an attention mechanism to extract a chapter's features, which ensures the consistency of labels [16]. However, the word2vec word vector from Zheng's method does not solve the problem of polysemy, which reduces the overall accuracy of recognition. In addition, the training of deep learning model usually requires a large number of manually labeled data, which can be a huge task in the fields of biology and medicine.

In order to cope with the above challenges of named entity recognition in biochemistry, this paper proposes a hybrid model (**B**ERT-**B**ILSTM-**M**HATT-**C**RF, BBMC) which mainly combines four approaches: **B**idirectional **E**ncoder **R**epresentations from **T**ransformers (BERT), **B**i-directional **L**ong **S**hort-**T**erm **M**emory (BILSTM), **M**ulti-**H**ead **Att**ention (MHATT), and **C**onditional **R**andom **F**ield (CRF). The main contributions of this paper are summarized as follows: (1) The above four key approaches are well integrated to address different challenging aspects of NER and the performance of this hybrid deep learning model is validated on a public-available dataset. (2) To learn high-level abstract information and address polysemy, the frequently-used word2vec model which produces word vector is replaced by a better representation model (BERT). (3) The multi-head attention mechanism in cognitive neuroscience is innovatively integrated to the BILSTM model to extract chapter-level features. (4) To improve the recognition rate, the softmax results of deep learning are connected to the CRF layer to make use of the dependency between tags.

The rest of the paper is organized as follows. Section 2 presents the BBMC model architecture and its four key modules: (1) BERT which has a large number of semantic features and part-of-speech features, (2) BILSTM which can predict the relationship between text sequences and tags, (3) MHATT that can extract information from multiple aspects of sentences, and (4) CRF which can predict the relationship between tags (because BERT can only predict the relationship between text sequence and label). The experimental datasets and setup of experiments are presented in Section 3. Section 4 demonstrates performance evaluation of the proposed hybrid model and comparison with existing approaches on the CHEMDNER dataset. The final section concludes our research efforts.

## 2. The hybrid BBMC model for biochemical named entity recognition

### 2.1. Architecture

The architecture of the proposed BBMC model is shown in Figure 1. The model consists of four parts: the BERT module, the BILSTM module, the MHATT module and the CRF module. The BERT module is first used to process the input data to obtain the underlying features of the input text, such as part-of-speech and characters, and primary sentence features. Next, the hidden layer of BILSTM stores two values that include forward and backward calculations. In this way, the target word can use the information at both the beginning and the end of the training process. Then, we apply the attention module to ensure that it can extract information of interest at the level of sentences. This module can address the problem of inconsistency, identify abbreviations and full-text labels (to a certain extent), and reduce the difficulty in training the model. Finally, the CRF layer decodes the output of the previous module into an optimal sequence and outputs the annotation information. The functions and underlying principles of each module are described in subsequent sections.

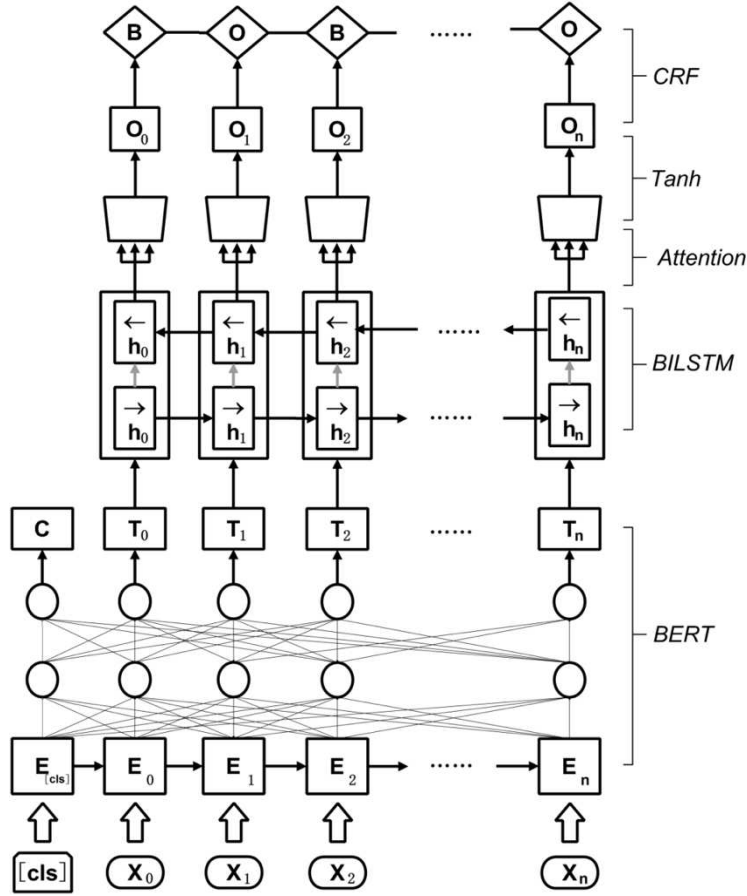

Figure 1. An overview of the BBMC model.

## 2.2. The BERT module

Currently, pre-training models are preferred in typical NLP tasks. The most prevalent one is word2vec, which was developed in an open source project by Google [22]. The maximum dictionary size of the word2vec's pre-training model can be in the millions, and the maximum dataset size of word2vec's pre-training model can reach 100 million. Once trained, via word embedding, the similarity between words can be measured, which is empirically efficient for most tasks. However, word2vec, as a static pre-training model, cannot address the problem of polysemy. There are two main strategies to apply pre-training models to NLP tasks. The first strategy is to use feature-based language models, such as **E**mbeddings from **L**anguage **Mo**dels (ELMO) [3]. The ELMO model introduces a contextual dynamic adjustment of word vectors to address the polysemy problem. However, the BILSTM structural feature extraction capability used is weaker than the Transformer we propose. The second strategy is to use the fine-tuning language model, such as the **G**enerative **P**re-trained **T**ransformer (GPT) [23]. This is a type of Transformer-based language model that can be adapted to multiple NLP tasks. It works by employing a pre-training language model for various tasks with very few (if any) changes to the model structure. However, it only uses a one-way language model, which limits its use in specific applications.

The BERT model that can perform multi-task learning based on a two-way deep network Transformer [24] is proposed. This model has created many advances in NLP and proved to be a good complement to the abovementioned models, alleviating their shortcomings. There are mainly two new component models proposed in BERT, the **M**asked **L**anguage **M**odel (MLM) and **N**ext **S**entence **P**rediction (NSP). The MLM randomly masks 15% of the words at each iteration, and the goal of the model is to predict these words given their context. Using transformer encoder with a strong extraction ability and using bidirectional training form, the MLM is ideally suitable for long sequence text named entity recognition tasks. The function of NSP is to understand relationships between sentences, and its specific role is to replace the sentence order of the corpus randomly, then predict the probability of the next sentence based on the previous sentence, and then continue to cycle training to get the results. In the training process, the loss functions of MLM and NSP are added to learn at the same time.

In this work, we adopt Google's open-source BERT pre-training model. The model input is a single sentence $X:=\{X_1,...,X_n\}$. BERT consists of three different layers, namely the Token layer, the Segment layer and the Position layer. The summed values of these layers are used as input to Transformer. The Token layer produces the word vector embedding associated with the word. The Segment layer distinguishes which sentence the word belongs to while the Position layer distinguishes the position of the word in the sequence. This process is represented in Figure 2.

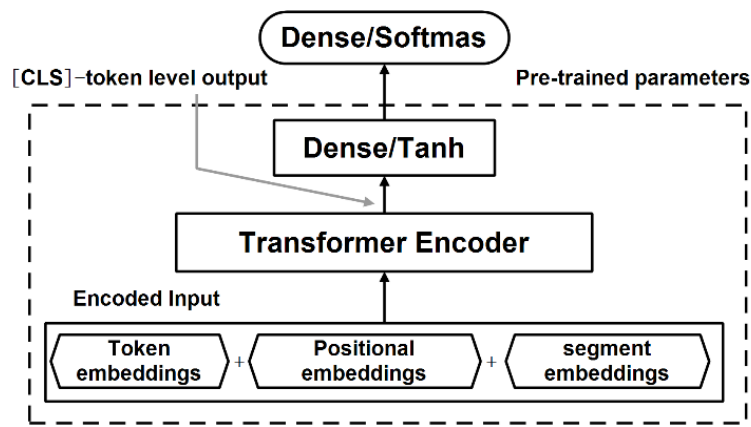

Figure 2. The structure of the Transformer.

After the processing provided by multiple Transformer modules, the BERT model is used to extract the features of the input sequence. Compared to the BILSTM, BERT processes the sequence step by step. Multiple transformer modules run in parallel to raise the speed of decoding and has strong capabilities for feature extraction.

The BERT model is excellent at feature extraction. It has been empirically verified that the fine-tuned BERT model has similar accuracy to the popular BILSTM-CRF model in NER task of chemicals and drugs. The main difference is that the latter can extract specific characteristics within a problem. However, the costs associated with the latter model is extremely high, such as a large amount of training time and high-level training difficulty. For these reasons, we combine the advantages of both models by adopting the BERT model instead of the traditional word2vec, and adopting a strategy of applying the BISLTM layer onto the BERT model.

## 2.3. The BILSTM module

The input text can be processed by the BERT module to directly output the corresponding recognition results according to the sentence probability vector. However, this method can yield relatively poor outcomes. A more effective approach is to only use the output of BERT as a representation of deep features such as part-of-speech, semantics, and primary sentence-level features. This output is then used as an input to the LSTM model to further extract context information.

LSTM, as a special case of RNN, is a cyclic RNN with long-term memory [25, 26]. Its network structure consists of one or more units with forgetting and memory functions, which overcomes the problem of traditional RNN gradient dispersion. This enables the network to retain the previous information selectively [27, 28]. In the case of biochemical named entity recognition, there are many cases where multiple words constitute an entity. With LSTM, the characteristics of long-distance dependence can be learned, thereby the ability of the model to identify long-sequence entities can be improved.

Let us denote the output sequence of the BERT model as $T:=\{T_1,...,T_n\}$. The corresponding LSTM diagram is shown in Figure 3.

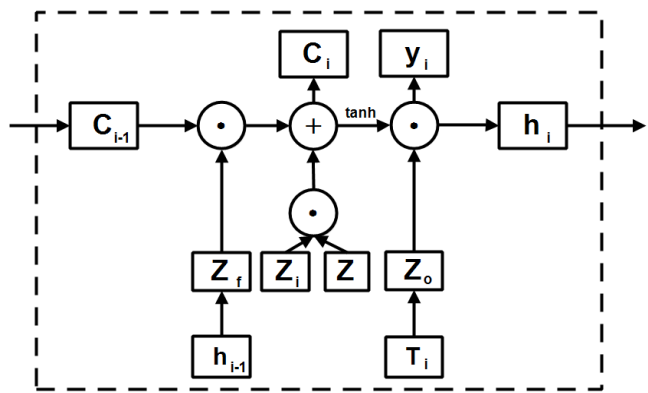

Figure 3．The internal structure of a LSTM/memory cell.

The LSTM has two transmission states: the cell state $C_i$ and the hidden state $h_i$. In comparison, the general RNN only has the single delivery state $h_i$. How the LSTM works can be formulated as follows,

$$Z_i = \sigma(W_i \cdot [T_i, h_{i-1}]) \qquad (1)$$

$$Z_f = \sigma(W_f \cdot [T_i, h_{i-1}]) \qquad (2)$$

$$Z_o = \sigma(W_{io} \cdot [T_i, h_{i-1}]) \qquad (3)$$

$$C_i = Z_f \odot C_{i-1} + Z_i \odot Z \qquad (4)$$

$$h_i = Z_o \odot \tanh(C_i) \qquad (5)$$

where $Z$, $Z_i$, $Z_f$ and $Z_o$ represent the i-th input unit status, input gate, forgetting gate, and output gate, respectively. $W_i$, $W_f$ and $W_0$ represent the $i$-th input gate, forgetting gate, and output gate weight matrix, respectively. Moreover, $h_i$ represents the i-th hidden layer state, $\odot$ represents Hadamard Product of matrix multiplication and $\sigma$ is the sigmoid function.

From the above equations, one can note that the hidden state $h_i$ of the LSTM only acquires information in the forward direction. We denote this using $\vec{h}_i$. BILSTM consists of LSTM modules for forward and reverse directions [7]. Compared with the unidirectional LSTM model, including the bidirectional output $[\vec{h}_i, \overleftarrow{h}_i]$ can better capture the two-way semantic dependence. Most current mainstream models use the advantages of BILSTM in the extraction of context information for feature extraction. In the domain of biochemically named entities, long sequence named entities are prevalent, however they are not well recognized using the BERT model. For this reason, we supplement the feature extraction of the BERT model and subsequently apply the BILSTM module to extract context information.

**2.4. The Multi-Head Attention module**

At the present, BILSTM is still the mainstream model of NER, which can memorize long text sequence features in theory. However, due to its gradient diffusion phenomenon, the application effect of BILSTM model needs to be improved [29]. In order to address this defect, attention mechanisms have been introduced into the field of computer vision [30]. The mechanisms can further enhance the memory ability and feature extraction ability of the BILSTM model for long sequence text, to a certain extent. They can also be used to address full-text label inconsistency and abbreviation recognition, and reduce the difficulty of model training.

As mentioned previously, in biochemical NER, there are a large number of abbreviated entities and long sequence named entities. For abbreviated entities, most authors use relevant abbreviations in articles after the first description of the relevant entities. Such articles usually have detailed descriptions of entities in earlier sections, and ordinary models can correctly identify corresponding entities based on, for example, context specific information. In other cases, the author uses abbreviations without introductions. Therefore, it is difficult for a conventional model to determine the correct entity label by contextual features. This may result in inconsistency in the labels of the entity throughout the entire article.

For long phrase NER, there are also many long-named entities (such as cyclic guanosine monophosphate) which exist in longer sentences. Such entities require stronger model processing capabilities including long sequence feature extraction. One possible solution is the current mainstream BILSTM-CRF model, where the CRF layer is labelled according to the features extracted by BILSTM. However, the model uses sentences as the basic processing unit and does not consider the features of the entire text. Moreover, in the processing of longer sequence texts, LSTM has been proven to have a feature extraction capability that is weaker than the model of attention mechanisms such as Transformer. Therefore, we adopt attention mechanisms to improve recognition accuracy.

As mentioned above, the BILSTM only focuses on modelling continuous context dependencies and ignores discrete context patterns. Though, discrete context correlation plays an important role in sequence labelling tasks. In general, for a given word, its label depends not only on its own semantic information and neighbouring context, but also on individual word information in the same sequence. This will greatly affect the accuracy of labelling. The study by Tan *et al.* proved that the self-attention mechanism can efficiently improve the performance on some NLP tasks, such as NER and POSl [31].

The Google mind team used the attention mechanism to classify images based on the RNN model [32]. Since then, the attention mechanism has been increasingly used in the field of computer vision. The Google machine translation team applied a large number of self-attention mechanisms to learn text representation and yielded excellent results[29].

Inspired by these recent advances, we apply the Multi-Head Attention mechanism to obtain relevant chapter-level features in biochemical NER from multiple angles and levels. By inputting the hidden layer output of the three settings of BILSTM, the information vector processed by the attention module is added by a certain weight to overcome the above disadvantages.

The Attention function takes a query and a set of key-value mappings as inputs, and outputs a weighted sum of values. The weight assigned to each value is calculated by the compatibility function of query and the corresponding key. The basis of the Multi-Head Attention used in this paper is the Scaled Dot-Product Attention mechanism, which uses the dot product to calculate the similarity between the query and each key[24]. The formula is given by

$$Atention(Q,K,V) = softmax(\frac{QK^T}{\sqrt{d_k}})V \qquad (6)$$

where $Q$, $K$ and $V$ represent matrix representations of query, key and value, respectively. The dot product is used to obtain the weight coefficient of the value corresponding to each key. Weights represent the importance of information and determine which values should be adjusted. The role of $\sqrt{d_k}$ is to control the dot product of Q and K, which is not so large as to avoid the problem of gradient disappearance after the processing by softmax. The process is shown in Figure 4.

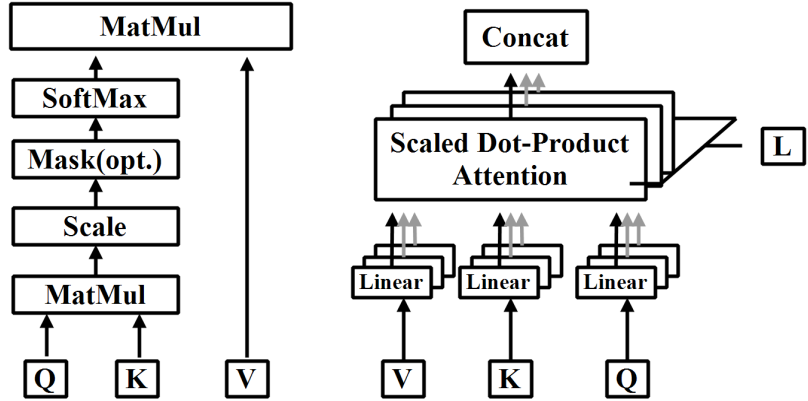

Figure 4. Structure of the Multi-Head Attention module.

Considering the necessity of extracting information from different representation subspaces at different positions, splicing many different Scaled Dot-Product Attentions result in a Multi-Head Attention. The specific process is to linearly project $h$ times of the queries $Q$, keys $K$ and values $V$ respectively.

$$head_i = Atention(QW_i^Q, KW_i^k, VW_i^v) \qquad (7)$$

$$MultiHead(Q,K,V) = Concat(head_i, ..., head_h) \qquad (8)$$

where $W_i^Q$, $W_i^K$ and $W_i^V$ represent the corresponding weight matrices of $Q$, $K$ and $V$, and the *Concat* function concatenates the results.

In summary, the distinguishing feature of the multi-head attention mechanism from the traditional attention mechanism is that it allows the model to learn relevant information in different representation subspaces. This feature can be well adapted to the recognition of the same entity in different contexts, especially the recognition of polysemy and abbreviations. Therefore, this paper integrates the Multi-Head attention mechanism into the model to improve the consistency of its full-text label recognition.

**2.5. The CRF module**

The current literature regards the NER as a sequence labelling task. We apply the **I**nside-**O**utside-**B**eginning (IOB) mechanism to mark the corpus. The structure of the annotation is shown in Table 1.

Table 1. The Structure of IOB.

| Entity tag | Start tag | Intermediate tag | End tag |
|---|---|---|---|
| TRIVIAL | B-TRIVIAL | I-TRIVIAL | E-TRIVIAL |
| FAMILY | B-FAMILY | I-FAMILY | E-FAMILY |
| … | … | … | … |
| NON-ENTITY | O | O | O |

Labels are not independent of each other, therefore, a key step is still to deal with the relationship between tags after obtaining the labelled dataset. One possibility is the CRF model, which is a conditional probability model that can be used to label ordered data. This model can handle the mutual constraint relationship between tags and solve the problem of sequence labelling effectively. Therefore, we apply the CRF model to processing the output of the Multi-head Attention module. The Viterbi algorithm is used to prediction and seeking the globally optimal annotation sequence[33].

We denote the sentence sequence as $x:=\{x_1,...,x_n\}$. After the processing via the above module, we obtain a $n \times m$ matrix $P$, where $n$ is the number of input words and $m$ is the number of label types. The entry $P_{ij}$ is the probability of the label $i$ of the word $j$ occurring in the sentence. We denote $y:=\{y_1,...y_i,...y_n\}$ as a tag sequence, thus the model computes the corresponding score:

$$Score(x,y) = \sum_{i-1}^{n} P_{i,y_i} + \sum_{i-1}^{n+1} A_{y_{i-1}y_i} \qquad (9)$$

where $A_{ij}$ is the transition probability from $y_i$ to $y_j$. We then apply softmax to obtain the normalized probability

$$P(y|x) = \frac{\exp(score(x,y))}{\sum_{y'} \exp(score(x,y'))} \quad (10)$$

For model training, the following maximized log likelihood function is used:

$$\log P(y^x|x) = score(x,y^x) - \log\left(\sum_{y'} \exp(score(x,y'))\right) \quad (11)$$

Finally, the model uses the Viterbi algorithm as follows to solve the optimal path in the prediction process.

$$y^* = \arg\max_{y'} score(x,y') \quad (12)$$

## 3. Experimental setup

### 3.1. Dataset

The dataset that we use is the CHEMDNER corpus, provided by the BioCreative project (https://biocreative.bioinformatics.udel.edu/news/biocreative-iii/), which is a community-driven project for the evaluation of NLP tasks in the biological domain [19]. It contains 84,435 manually labelled chemical entities. The details of the data setting are shown in Table 2.

Table 2. Main entity categories and their numbers in the dataset.

| Category | Training setting | Development setting | Test setting | Entire corpus |
|---|---|---|---|---|
| Abstracts | 3,500 | 3,500 | 3,000 | 10,000 |
| Abstracts with CEM | 2,916 | 2,907 | 2,478 | 8,301 |
| Nr.journals | 193 | 188 | 188 | 203 |
| Nr.chemicals | 8,520 | 8,677 | 7,563 | 19,805 |
| Nr.Mentions | 29,478 | 29,526 | 25,351 | 84,355 |
| TRIVIAL | 8,832 | 8,970 | 7,808 | 25,610 |
| SYSTEMATIC | 6,656 | 6,816 | 5,666 | 19,138 |
| ABBREVIATION | 4,538 | 4,521 | 4059 | 13,118 |
| FORMULA | 4,448 | 4,137 | 3,443 | 12,028 |
| FAMILY | 4,090 | 4,223 | 3,622 | 11,935 |
| IDENTIFIER | 672 | 639 | 513 | 1,824 |
| MULTIPLE | 202 | 188 | 199 | 589 |
| NO CLASS | 40 | 32 | 41 | 113 |

It can be seen that the proportions of different types of entities in the text are different, and some entities appear less frequently in the dataset, therefore we call them as low-frequency entities. The recognition accuracy of a low-frequency entity is relatively poor because the number of samples is small and most of them are special entities.

### 3.2. Parameter settings

Although the BERT model can be fine-tuned, its optimization performance is limited. This hybrid model integrates BILSTM, MHATT and CRF; therefore, the hyperparameters need to be optimised. The model hyperparameters are presented in Table 3. Among them, the weight attenuation coefficient (weight_decay) is used to prevent over-fitting. The purpose of gradient cropping coefficient (clip) is to prevent gradient explosion.

Table 3. The hyperparameters and their values used in the BBMC model.

| Parameters | Description | Value |
| --- | --- | --- |
| lr | Initial learning rate | 0.001 |
| dropout | Loss rate | 0.5 |
| embedding_dim | Embedded layer dimension | 768 |
| enc_hidden_dim | Coding layer dimension | 256 |
| lstm_dim | Single layer LSTM dimension | 128 |
| key_dim | The size of $K$ in the Attention module | 64 |
| val_dim | The size of $V$ in the Attention module | 64 |
| num_heads | The size of $L$ in the Attention module | 3 |
| weight_decay | Weight attenuation coefficient | 0.01 |
| clip | Gradient cropping | 5 |

### 3.4. Performance comparison

To evaluate the performance of the BBMC model, we compare its performance with that of five state-of-the-art models: BILSTM-CRF (BC), BERT-BILSTM-CRF (BBC), ELMO-BILSTM-CRF (EBC), BILSTM-MULATT-CRF (BMC), and ELMO-BILSTM-MULATT-CRF (EBMC).

The bi-directional structure of BC can obtain the sequence information of context, therefore, it is widely used in tasks such as named entity recognition [34]. BERT in BBC can pre-train dynamic word vector which can express different semantics in different contexts [24]. ELMO in EBC, as a deep contextual word representation, can model complex features of words [35]. Attention mechanism is added to BC model to avoid complex feature engineering in traditional work [34]. The purpose of MULATT's long attention mechanism in BMC is to replenish the key information of sequence from multiple aspects [29].

It is noted that BC, BBC, and EBC do not use the attention mechanism; BC and BMC use the GloVe word vector while BBC and EBMC use the ELMO pre-training model; EBC and BBMC use the BERT pre-training model. To ensure the fairness, the hyper parameters of all models are consistent.

In order to evaluate the model performance, the accuracy, recall and F-score are used as the evaluation metrics.

## 4. Experimental results and discussion

### 4.1. Abbreviations and polysemy labels

In order to explain the inconsistency of full-text labelling and polysemy labelling mentioned above, this paper analyses different models for identifying special entities. Due to the space limitation, we illustrate with only two examples, as shown in Table 4.

Table 4. Performance comparison of different entity identification methods (Left: effects recognized by the proposed BBMC model. Right: Effects recognized by BILSTM -CRF)

| Model | BBMC | BILSTM-CRF |
|---|---|---|
| **Abbreviation** | The fabrication of *patterned microstructures in poly(dimethylsiloxane) (***PDMS**) is a prerequisite for soft lithography . Herein , curvilinear surface relief microstructures in **PDMS** are fabricated through a simple three - stage approach combining microcontact printing ( CP ) , selective surface wetting / dewetting and replica molding ( REM ) . First , using an original **PDMS** stamp (first - generation stamp) with linear relief features , ……Finally , based on a REM process , the PEG -dot array on gold substrate is used to fabricate a second - generation **PDMS** stamp with microcavity array , and the second - generation **PDMS** stamp is used to generate third - generation **PDMS** stamp with microbump array . | The fabrication of *patterned microstructures in poly(dimethylsiloxane)*(PDMS) is a prerequisite for soft lithography . Herein , curvilinear surface relief microstructures in PDMS are fabricated through a simple three - stage approach combining microcontact printing ( CP ) , selective surface wetting / dewetting and replica molding ( REM ) . First , using an original PDMS stamp (first - generation stamp) with linear relief features , ……Finally , based on a REM process , the PEG -dot array on gold substrate is used to fabricate a second - generation PDMS stamp with microcavity array , and the second - generation PDMS stamp is used to generate third - generation PDMS stamp with microbump array . |

| | Silver was found to be only active in form of <u>free **silver ions**</u> ( FSI ) . | Silver was found to be only active in form of <u>free silver ions</u> ( FSI ) . |
| --- | --- | --- |
| **Polysemy** | …… | …… |
| | High glucose insulin and **<u>free fatty acid</u>** concentrations synergistically enhance perilipin 3 expression and lipid accumulation in macrophages . | High glucose insulin and free fatty acid concentrations synergistically enhance perilipin 3 expression and lipid accumulation in macrophages . |

Table 4 mainly marks the identification of two individual words: **PDMS** and **free**. The **PDMS** is an abbreviation of patterned microstructures in poly (dimethylsiloxane). The words in bold and underlined in Table 4, meaning to be correctly identified as entities. Table 4 shows that the BBMC model recognizes both the full name and the abbreviation. In contrast, the BILSTM -CRF model only correctly identifies the full name. The reason is that BBMC model introduces an attention mechanism to extract chapter-level information, which ensures the label consistency of the full text. The individual word *Free* is a polysemous word. For example, in the phrase of *free silver ions*, the *free* word is not an entity and is used to modify the entity silver ions. The phrase of *free fatty acid, as* a fixed phrase, can be identified by the BBMC model. In other words, the BBMC model can recognize polysemous words in different situations. However, the **BILSTM-CRF** model is relatively ineffective in recognizing polysemous words due to lack of the representation ability of sentence features. This also proves that the BERT model can better adjust the word vector representation according to the context compared to pre-training models such as GloVe [36].

**4.2. Visual analysis of the BERT model**

Instead of using the traditional word vector as the input, the BERT model uses the attention-based transformer structure. This section reports the analysis results about the attention distribution of BERT in NER. For better demonstration, this paper selects "*systematic has been used for thousands of years, although systematic is clearly toxic to most mammalian organ systems*" as an input text and three experimental examples are shown in Figures 5(a)-(c). At the top of the figure, the user can select one or more attention heads that are represented by the coloured squares. The word systematic in the picture is the entity to be recognized. It can be seen in Figure 5(a) that the second "systematic" on the right pays more attention to "systematic" and "although" in front of it and "is" behind it. From Figure 5(b), the first "systematic" on the right has a high degree of attention to the "has", "years" and "systematic". Therefore, the subsequent model can determine the annotation type of the word by the part-of-speech and the relationship between the words before and after. Moreover, the attention mechanism of cross sentence can also effectively improve the adaptability of subsequent models to long sentences. In Figure 5(c), the word "although", as a conjunction, focuses on the separators, that is, the relationship between sentences. This special effect is very important for NLP tasks, but for NER tasks, it results in active weakening of focusing on such words. To sum up, the BERT model has a stronger tendency of attention to meaningful entities to be recognized, while weakening the attention of unimportant words enhances the training efficiency and accuracy of the model.

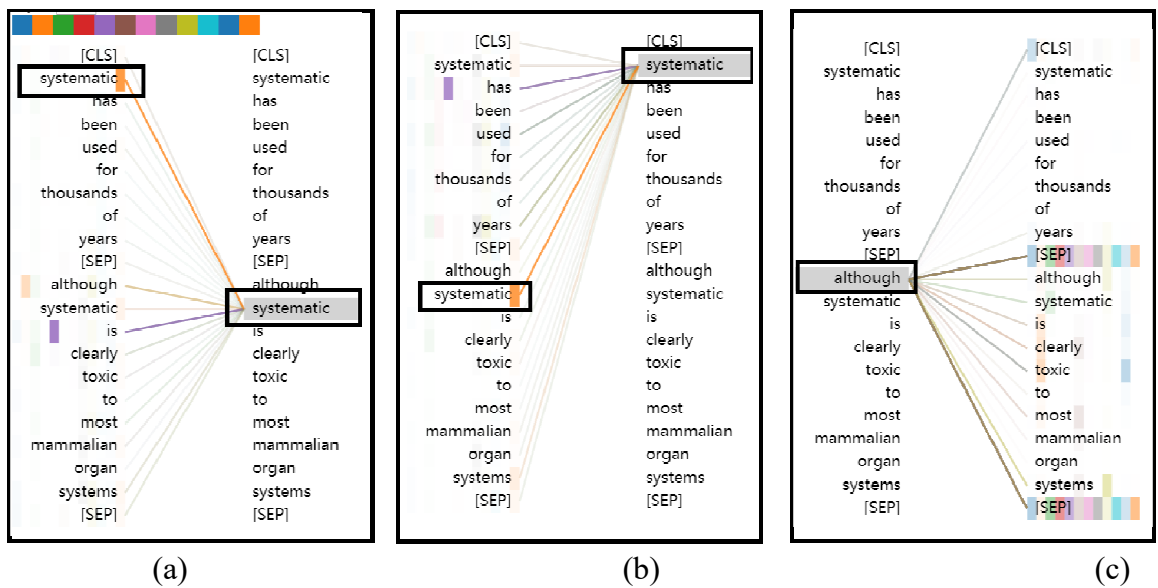

(a) (b) (c)

Figure 5. Examples of attention-head views produced from BBMC. (a) and (b) demonstrate different attention-head views of the word "systematic" in different positions, and (c) depicts attention-head view for the conjunction of "although". Self-attention is represented as lines connecting the tokens that are showing solicitude for (left) with the tokens being attended to (right). Colours sign the appropriate attention head(s), while line weight mirrors the attention level.

**4.3. Low-frequency and high-frequency entity recognition**

Low-frequency word entity recognition is more challenging than high-frequency word entity recognition. Figure 6 shows the numbers of seven different entity datasets (MULTIPLE, IDENTIFIER, FAMILY, FORMULA, ABBREVIATION, SYSTEMATIC, and TRIVIAL; see Section 3.1 for more details of the entity datasets) and the recognition accuracies (represented by F-value) for these entities by six different models (BC, BBC, EBC, BMC, EBMC, BBMC, as shown in Table 5).

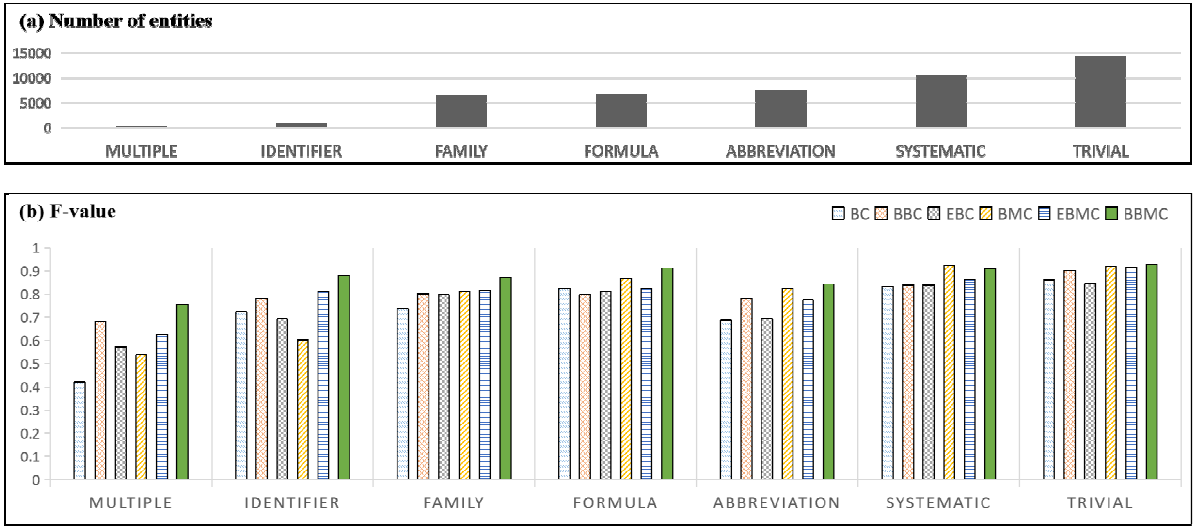

Figure 6. The numbers of seven different categories of entities (Figure 6(a)) and the recognition accuracies (represented by F-value) for seven entity categories achieved by six different models (Figure 6(b)). BC: BILSTM-CRF, BBC: BERT-BILSTM-CRF, EBC: ELMO-BILSTM-CRF, BMC: BILSTM-MULATT-CRF, EBMC: ELMO-BILSTM-MULATT-CRF, and BBMC: BERT-BILSTM-MULATT-CRF.

Figure 6 demonstrates that the six models achieve higher accuracies in recognising high-frequency words (such as SYSTEMTIC and TRIVIAL datasets). However, there is still room for improving the recognition accuracy in the case of low-frequency words (such as MULTIPLE and IDENTIFIER datasets). The performance of BBMC and BBC shows improvements in the accuracy of recognizing low-frequency entities. This suggests the large-scale pre-training retention of the BERT model is able to supplement scarce training data. BBMC outperforms the other five models in identifying low-frequency entities. For example, compared with the recognition accuracies for low-frequency entities produced by BC in MULTIPLE and IDENTIFIER datasets, those produced by BBMC are increased by 80% and 21.69%, respectively. Overall, for low-frequency, middle-frequency (entities in FAMILY, FORMULA, and ABBREVIATION datasets), and high-frequency words, BBMC nearly achieves the best recognition performance.

**4.4. Performance comparison with existing models**

Three key findings can be drawn from the results in Table 5. Firstly, BMC, EBMC, and BBMC can achieve better performance by using the attention mechanism. This demonstrates that the attention mechanism can eliminate data noise and combine chapter-level information to improve recognition accuracy. Secondly, a model can achieve a higher F-value by adding BERT or ELMO than the models adding GloVe. This demonstrates that dynamic adjustment of word vectors can improve the accuracy of polysemy recognition. Thirdly, a model which uses BERT will yield slight improvement over a model which uses ELMO. This is mainly due to the fact that the Transformer used by BERT is more powerful than the BILSTM feature extraction used by ELMO.

Table 5. Model performance comparison.

| Abbreviation | Model | P (%) | R (%) | F (%) |
|---|---|---|---|---|
| BC | BILSTM-CRF | 91.31 | 87.73 | 89.48 |
| BBC | BERT-BILSTM-CRF | 89.9 | 88.6 | 89.2 |
| EBC | ELMO-BILSTM-CRF | 89.2 | 88.9 | 89.04 |
| BMC | BILSTM-MULATT-CRF | 91.4 | **90.15** | 90.77 |
| EBMC | ELMO-BILSTM-MULATT-CRF | 90.9 | 89.05 | 89.96 |
| BBMC | BERT-BILSTM-MULATT-CRF | **91.8** | 89.9 | **90.84** |

The evaluation results show that the proposed hybrid model outperform existing models in the literature. The outstanding performance is achieved due to the following reasons. First, the BERT model is introduced in this paper as a component/module, compared to the traditional use of the word2vec and GloVe, is capable of adjusting the vector of the target word according to the context, to some extent, and account for word polysemy. The Transformer is introduced in relation to the word vector, which provides strong feature extraction capabilities. At the same time, the features retained by the large-scale pre-training model improve the recognition rate of entities in categories which suffer from data scarcity. Second, we use Multi-Head Attention in addition to BILSTM to process specific information. The use of chapter-level information reduces the occurrence of inconsistency in full-text entity tags. This also improves the recognition ability of the model for long sequence sentences. Third, in order to avoid the final use of softmax to calculate the labelling results, we apply the CRF layer. The use of dependencies between tags improves the recognition rate.

## 5. Conclusion

We propose a new hybrid deep-learning model for NER in biochemistry. The model is underpinned by innovatively integrating BERT, BILSTM, MHATT, and CRF organically. Through experiments and comparative analyses, we have shown that the new model has outstanding performance in identifying abbreviations, polysemous words, both low-frequency and high-frequency words, as well as ensuring the consistency of full-text labels. This paper confirms the effectiveness of the multi-head attention mechanism in the field of natural language processing to improve the performance of models. At the same time, this paper further verifies that Transformer's ability to extract features is better than BILSTM, and Transformer has lower training difficulty. Therefore, our future work will further make greater use of the Transformer instead of BILSTM, in order to verify the advantages and stability of Transformer compared to BILSTM through different types of text and application fields, with the aim of achieving higher recognition accuracy.

In addition, we have noticed in our research that when further performing the task of relationship extraction, the accuracy difference of NER is further enhanced in the extracted results. This means that the error of the model may further propagate, and hence expand in downstream tasks. To this end, the joint training of the model and relationship extraction in this paper will effectively reduce the propagation error.

**Acknowledgments**

This work was supported in part by the National Natural Science Foundation of China (No. 61906057, 61703306), the Fundamental Research Funds for the Central Universities, the Fundamental Research Funds for the Central Universities of China, and Natural Science Foundation of Anhui Province.